\documentclass[runningheads]{llncs}
\usepackage{graphicx}

\usepackage{times}
\usepackage{latexsym}
\usepackage{url}
\usepackage{tabularx}
\usepackage{booktabs}
\usepackage[utf8]{inputenc}
\usepackage[T1]{fontenc}
\usepackage{CJKutf8}
\usepackage{xcolor}
\usepackage{subfig}
\usepackage{multirow}
\usepackage{array}
\newcolumntype{?}{!{\vrule width 1.5pt}}
\newcolumntype{C}[1]{>{\centering\let\newline\\\arraybackslash\hspace{0pt}}m{#1}}
\newcolumntype{L}[1]{>{\raggedright\let\newline\\\arraybackslash\hspace{0pt}}m{#1}}
\usepackage{enumitem}

\usepackage{lastpage}
\usepackage{fancyhdr} 
\usepackage[acronym,nomain,nonumberlist]{glossaries}
\newacronym{s2s}{Seq2Seq}{Sequence-to-Sequence}
\newacronym{nlp}{NLP}{Natural Language Processing}
\newacronym{nmt}{NMT}{Neural Machine Translation}
\newacronym{cnn}{CNN}{Convolutional Neural Network}
\newacronym{dl}{DL}{Deep Learning}
\newacronym{mt}{MT}{Machine Translation}
\newacronym{bpe}{BPE}{Byte-Pair Encoding}
\newacronym{gru}{GRU}{Gated Recurrent Unit}
\newacronym{rnn}{RNN}{Recurrent Neural Network}
\newacronym{sgd}{SGD}{Stochastic Gradient Descent}
\newacronym{map}{MAP}{Maximum A Posteriori}
\newacronym{ml}{ML}{Maximum Likelihood}
\newacronym{pos}{PoS}{part-of-speech}
\newacronym{dnn}{DNN}{Deep Neural Network}
\newacronym{mlp}{MLP}{Multi-layer Perceptron}

\makeatletter
\newcommand{\printfnsymbol}[1]{%
  \textsuperscript{\@fnsymbol{#1}}%
}
\makeatother

\begin{document}

\title{Towards Recognizing Phrase Translation Processes: Experiments on English-French}



\author{Yuming Zhai\thanks{Both authors contributed equally to this article.} \and
Pooyan Safari\printfnsymbol{1} \and
Gabriel Illouz \and
Alexandre Allauzen \and
Anne Vilnat 
}
%
%
\institute{LIMSI-CNRS, Univ. Paris-Sud, Univ. Paris-Saclay, France \\ 
\email{\{firstname.lastname\}@limsi.fr}}

\maketitle              


\vspace{-2em}

\begin{abstract}
When translating phrases (words or group of words), human translators, consciously or not, resort to different translation processes apart from the literal translation, such as Idiom Equivalence, Generalization, Particularization, Semantic Modulation, \textit{etc.} Translators and linguists (such as Vinay and Darbelnet, Newmark, \textit{etc.}) have proposed several typologies to characterize the different translation processes. However, to the best of our knowledge, there has not been effort to automatically classify these fine-grained translation processes. Recently, an English-French parallel corpus of TED Talks has been manually annotated with translation process categories, along with established annotation guidelines. Based on these annotated examples, we propose an automatic classification of translation processes at subsentential level. Experimental results show that we can distinguish non-literal translation from literal translation with an accuracy of 87.09\%, and 55.20\% for classifying among five non-literal translation processes. This work demonstrates that it is possible to automatically classify translation processes. Even with a small amount of annotated examples, our experiments show the directions that we can follow in future work. One of our long term objectives is leveraging this automatic classification to better control paraphrase extraction from bilingual parallel corpora. 


\end{abstract}


\vspace{-2.5em}

\section{Introduction}
\label{introduction}

Since 1958, translators and linguists have published work on translation processes~\cite{VinayDarbelnet1958,Newmark1981,ChuquetPaillard1989,Molina_HurtadoAlbir_2002}.
They distinguish literal translations from other translation processes at subsentential level. Consider these two human non-literal translation examples: 
the first translation preserves exactly the meaning,
where the fixed expression \textit{à la hauteur de}~`to the height of' has a figurative sense which means \textit{capable of solving}; while the second one is more complicated, there exists a textual inference between the source text and the translation. 

\vspace*{0.5em}

\noindent \textit{(1.EN) a solution \textbf{that's big enough to solve} our problems} \\
\textit{(1.FR) une solution \textbf{à la hauteur de} nos problèmes}  \\
\textit{(2.EN) and \textbf{that scar has stayed with him} for his entire life} \\
\textit{(2.FR) et que, toute sa vie, \textbf{il a souffert de ce traumatisme}} \\
(`he has suffered from this traumatism') 

\vspace*{0.5em}

Non-literal translations can bring difficulties for automatic word alignment~\cite{Dorr_DUSTer_2002,Deng2017}, or cause meaning changes in certain cases. However, to the best of our knowledge, there has not been effort to automatically classify these fine-grained translation processes to benefit downstream natural language processing tasks. For example, \gls{mt} techniques have been leveraged for paraphrase extraction from bilingual parallel corpora~\cite{BannardCallison_Burch2005,Mallinson2017}. The assumption is that two monolingual segments are potential paraphrases if they share common translations in another language. Currently the largest paraphrase resource, PPDB (ParaPhrase DataBase)~\cite{Ganitkevitch2013}, has been built following this method. Nonetheless, Pavlick \textit{et al.}~\cite{PPDB_relation2015} revealed that there exist other relations (\textit{i.e. Entailment (in two directions), Exclusion, Other related and Independent})\footnote{Exclusion: X is the contrary of Y; X is mutually exclusive with Y. Other related: X is related in some other way to Y. (\textit{e.g. country / patriotic}). Independent: X is not related to Y.} than strict equivalence (paraphrase) in PPDB. Non-literal pivot translations inside the parallel corpora could break the strict equivalence between the candidate paraphrases extracted, whereas they have not received enough attention during this corpora exploitation.

From a linguistic point of view, apart from the word-for-word literal translation, different versions of human translations reflect the richness of human language expressions, where various translation processes could be employed. Furthermore, because of the existing differences between languages and cultures, non-literal translation processes are sometimes inevitable to produce correct and natural translations. The fine-grained phrase-level translation processes could help foreign language learners to better compare the language being learned with 
another language already mastered. 
 
Based on the theories developed in translation studies and through manually annotating and analyzing an English-French parallel corpus, Zhai \textit{et al.}~\cite{zhai2018b} have proposed a typology of translation processes adapted to their corpus. In this work our main contribution is proposing an automatic classification of translation processes at subsentential level, based on these annotated examples. From the aspect of granularity and our goal of better controlling paraphrasing process or helping foreign language learners, it is different from the task of filtering semantically divergent parallel sentence pairs to improve the performance of \gls{mt} systems~\cite{Carpuat2017,Vyas2018,Pham2018}. Experimental results show that we can distinguish non-literal translation processes from literal translation with an accuracy of 87.09\%, and 55.20\% for classifying among non-literal multi-classes. 

In the present paper, after reviewing related work, we describe the manual annotation and the data set. Exploited features and different neural network architectures will be presented, followed by experimental results and error analysis. Finally we conclude and present the perspectives of this work.


\section{Related Work}
\label{relatedWork}

Translators and linguists have proposed several typologies to characterize different translation processes. Vinay and Darbelnet~\cite{VinayDarbelnet1958} identified direct and oblique translation processes, the latter being employed when a literal translation is unacceptable, or when structural or conceptual asymmetries arising between the source language and the target language are non-negligible. Following studies include, among others, the work of Newmark~\cite{Newmark1981,newmark1988}, Chuquet and Paillard \cite{ChuquetPaillard1989}. More recently, Molina and Hurtado Albir~\cite{Molina_HurtadoAlbir_2002} proposed their own categorization based on studying the translation of cultural elements in the novel \textit{A Hundred Years of Solitude} from Spanish to Arabic. 

Non-literal translations or cross-language divergences have been studied to improve \gls{mt} related techniques. In order to enable more accurate word-level alignment, Dorr \textit{et al.}~\cite{Dorr_DUSTer_2002} proposed to transform English sentence structure to more closely resemble another language. A translation literalness measure was proposed to select appropriate sentences or phrases for automatically constructing \gls{mt} knowledge~\cite{imamura2003automatic}. Using a hierarchically aligned parallel treebank, Deng and Xue~\cite{Deng2017} semi-automatically identify, categorize and quantify seven types of translation divergences between Chinese and English.\footnote{Lexical encoding; difference in transitivity; absence of language-specific function words; difference in phrase types; difference in word order; dropped elements; structural paraphrases.} Based on the syntactic and semantic similarity between bilingual sentences, Carl and Schaeffer~\cite{Carl_Schaeffer_2017} developed a metric of translation literality. We have drawn inspiration from these preceding work for our feature engineering. 

Recently, different models have been proposed to automatically detect translation divergence in parallel corpora, with the goal of automatically filtering out divergent sentence pairs to improve \gls{mt} systems' performance. An SVM-based cross-lingual divergence detector was introduced \cite{Carpuat2017}, using word alignments and sentence length features. Their following work~\cite{Vyas2018} proposed a Deep Neural Network-based approach. This system could be trained for any parallel corpus without any manual annotation. They confirmed that these divergences are a source of performance degradation in neural machine translation. Pham \textit{et al.}~\cite{Pham2018} built cross-lingual sentence embeddings according to the word similarity with a neural architecture in an unsupervised way. They measure the semantic equivalence of a sentence pair to decide whether to filter it out. 

Another task studying human translations concerns automatic post-editing~\cite{APE_2018}. The aim is evaluating systems for automatically correcting translation errors of an unknown ``black box'' \gls{mt} engine, by learning from human revisions of translations produced by the same engine. Evaluation metrics include TER~\cite{snover2006study}, BLEU~\cite{papineni2002bleu} and manual evaluation. The task that we propose here is different from these attempts, which either filter semantically divergent sentence pairs to improve the performance of \gls{mt} systems; or automatically correct machine translation errors to improve the translation quality. Our task of classifying translation processes (in two classes or in multi-classes) at subsentential level is a stand-alone task. One of our long term objectives is leveraging this automatic classification to better control phrase-level paraphrase extraction from bilingual parallel corpora. 


\section{Manual Annotation and Data Description}
\label{annotation}

In order to model translation choices made by human translators at subsentential level, Zhai \textit{et al.}~\cite{zhai2018b} have annotated a trilingual parallel (English-French, English-Chinese) corpus of TED Talks\footnote{https://www.ted.com/} with translation processes. The corpus is composed of transcriptions and human translations of oral presentations. The inter-annotator agreement (Cohen's Kappa)~\cite{Cohen1960} for annotating the English-French and English-Chinese control corpus is 0.67 and 0.61, both around the substantial agreement threshold. This indicates that the task of manual annotation is already complicated. Readers can find more details of corpus construction in the article~\cite{zhai2018b}. 

The automatic classification is conducted on the English-French pair in this work. We present in the table~\ref{tab:definition} a brief definition, a typical example and the number of instances for each category to be automatically classified.\footnote{Note that there are other detailed annotation rules in the annotation guidelines.} We combine \textit{Transposition} and \textit{Mod+Trans} in a category \textit{Contain\_Transposition}, where \textit{Modulation} is considered as a neutral part. We will work on the classification of the pair English-Chinese once the annotation phase is finished. In this work, we conduct experiments in a simplified scenario, where we already know the boundaries of bilingual pairs, and we only predict the translation process. For example, given the pair \textit{deceptive $\rightarrow$ une illusion} in a pair of bilingual sentences, the goal is to predict its label \textit{Contain\_Transposition}. 

\vspace*{-1em}

\begin{table}[!ht]
  \centering 
    \caption{Definition, typical example and number of instances for each translation process to be automatically classified. The instances were manually annotated in an English-French parallel corpus of TED Talks. We combine \textit{Transposition} and \textit{Mod+Trans} in a category \textit{Contain\_Transposition} for the automatic classification.}
    \begin{tabularx}{\textwidth}{l|L{9.5cm}}
    \toprule
    Translation Process 
    & Definition and typical example \\ \midrule
    Literal & Word-for-word translation, also concerns lexical units in multiword form.  \\ 
    (3771) & \textit{certain \textbf{kinds} of $\rightarrow$ certains \textbf{types} de} \\ \hline 
    
    Equivalence & Non-literal translation of proverbs or fixed expressions; a word-for-word translation makes sense but the translator expresses differently, without changing the meaning and the grammatical classes. \\
    (289) & \textit{\textbf{back then} $\rightarrow$ \textbf{à l'époque} (`at that time')} \\ \hline
    
    Generalization & Several source words or expressions could be translated into a more general target word or expression, the translator uses the latter to translate.  \\ 
    (86) & \textit{as we \textbf{sit here} in ... $\rightarrow$ alors que nous \textbf{sommes} à ... (`as we are at ...')} \\  \hline
    
    Particularization & The source word or expression could be translated into several target words or expressions with a more specific meaning, and the translator chooses one of them according to the context. \\ 
    (215) & \textit{the idea I want to \textbf{put out} is ... $\rightarrow$ l'idée que je veux \textbf{diffuser} c'est ... (`the idea I want to spread is ...')} \\ \hline
    
    Modulation & Metonymical and grammatical modulation~\cite{ChuquetPaillard1989}; change the point of view; the meaning could be changed. \\
    (195) & \textit{\textbf{that scar has stayed with him} $\rightarrow$ \textbf{il a souffert de ce traumatisme} (`he has suffered from this traumatism')} \\ \hline
    
    Transposition & Change grammatical classes without changing the meaning. \\
    (289) & \textit{unless \textbf{something changes} $\rightarrow$ à moins qu'\textbf{un changement ait lieu} (`unless a change occurs')} \\ \hline
    
    Mod+Trans & Combine the transformations of \textit{Modulation} and of \textit{Transposition}, which could make the alignment difficult. \\ 
    (53) & \textit{this is a completely \textbf{unsustainable} pattern $\rightarrow$ il est absolument \textbf{impossible de continuer sur} cette tendance (`it is completely impossible to continue on this trend')} \\ 
    \bottomrule 
    \end{tabularx}
    \label{tab:definition}
\end{table}


\section{Automatic Classification}
\label{classification}

We have tried two approaches for the automatic classification. Since the size of the cross validation data set is quite small, we first compare different statistical machine learning techniques with feature engineering. We also build different neural network architectures which we explain below. 

\subsection{Feature Engineering with Statistical Machine Learning Techniques}

We describe below the features exploited in this work. The tag sets of English and French for \gls{pos} tagging, constituency parsing and dependency parsing have been converted into three compact and unified tag sets~\cite{UniversalPosTag2012}.  

1) The \gls{pos} tagging is done by \textit{Stanford CoreNLP}~\cite{stanford-coreNLP2014} for the two languages. 
On source and target side, for each \gls{pos} tag, the number of its occurrence is counted in a vector. 
We also calculate the cosine similarity between these two vectors (on all words and only on content words).\footnote{The tags of content words include: ADJ, ADV, NOUN, PROPN, VERB. If a segment does not contain any content word, the original segment is used.}

2) We verify the pattern of \gls{pos} tag sequence changing according to a manual list, for example the pair \textit{methodologically $\rightarrow$ de façon méthodologique} `methodologically' corresponds to the pattern \textit{ADV $\rightarrow$ ADP NOUN ADJ}.  

3) The number of tokens in the two segments ($l_e$, $l_f$), the ratio of these numbers ($l_e/l_f$, $l_f/l_e$), the distance Levenshtein~\cite{Levenshtein1966} between the segments. 

4) The constituency parsing is done by \textit{Bonsai}~\cite{Bonsai2010} for French, by \textit{Stanford CoreNLP} for English. We compare the \gls{pos} tags for a pair of words, the non-terminal node tags for a pair of segments, the tag category (\textit{e.g.} verb $\rightarrow$ verb phrase) for a word translated by a segment or vice versa. 

5) The dependency parsing is done by \textit{Stanford CoreNLP} for the two languages. Inside the segments, the number of occurrence of each dependency relation is counted. Outside the segments, among the words linked at source and target side, we filter those which are aligned in the sentence context. Then the number of occurrence of each dependency relation between the words in segments and these context words is counted.  

6) The cosine similarity is calculated between the embeddings from \textit{ConceptNet Numberbatch}~\cite{conceptNet5_5_2017}. 
This resource is multilingual and the system based on \textit{ConceptNet} took the first place in the task ``Multilingual and Cross-lingual Semantic Word Similarity'' of SemEval2017~\cite{Semeval2017_task2,ConceptNet_semeval2017}. 
Certain multi-word expressions have their own embeddings in this resource. Otherwise, we calculate the average of embeddings only on content words. The same features are calculated for lemmatized segments.\footnote{The lemmatization is done by \textit{Stanford CoreNLP} and \textit{Tree Tagger}~\cite{Treetagger1995} for English and French.} 

7) The resource \textit{ConceptNet}~\cite{conceptNet5_5_2017} also provides assertions in triplet: a pair of words or expressions linked by a relation.
In this multilingual resource, we verify if an English-French pair is directly linked; indirectly linked by another French segment or simply not linked.\footnote{The EN-FR and FR-FR assertions are used in this work.} Three forms are tested: original form, lemmatized form and lemmatized filtered form.\footnote{We filter the words in a manual list, for example the light verbs, determinants, pronouns, \textit{etc.}} 

8) On the lemmatized filtered form, we calculate the percentage of tokens which are linked with a relation of derivation, based on the resource \textit{ConceptNet}. For example \textit{deceptive} and \textit{illusion} `illusion' are not directly linked in the resource, but they are both linked to \textit{illusoire} `illusory'. Hence we consider that there exists a link of derivation between them.
\\

For the three following features, we have exploited the lexical translation probability table generated by the statistical word alignment tool \textit{Berkeley Word Aligner}~\cite{BerkeleyAligner2006}, trained on an English-French parallel corpus composed of TED Talks and a part of Paracrawl corpus (in total 1.8M parallel sentence pairs and 41M English tokens).\footnote{https://wit3.fbk.eu/, https://paracrawl.eu/index.html}

9) The entropy of the distributions of lexical translation probabilities~\cite{Entropy1990,Carl_Schaeffer_2017}, calculated according to this equation: $H(X) = \sum_{i}P(x_i)I(x_i) = -\sum_{i}P(x_i)log_eP(x_i)$. We calculate the average entropy on content words. A bigger entropy indicates that the words have more general meanings or they are polysemous. The same feature is calculated on the lemmatized content words.  

10) The bidirectional lexical weighting on content words, by supposing a \textit{n-m} alignment $a$ between the segments ($\bar{e}$ and $\bar{f}$). In the scheme proposed by Koehn \textit{et al.}~\cite{lexicalWeighting2003} (equation~\ref{eq:lexWeighting}), to calculate the direct lexical weighting, each of the English words $e_i$ is generated by aligned foreign words $f_j$ with the word translation probability $w(e_i|f_j)$. And similarly for the reverse lexical weighting $lex(\bar{f}|\bar{e},a)$. The same feature is calculated for lemmatized content words. This feature could reflect the alignment confidence between a pair of segments. 

\begin{equation}
    \label{eq:lexWeighting}
    lex(\bar{e}|\bar{f}, a) = \prod_{i=1}^{length(\bar{e})} \frac{1}{|\{j|(i,j) \in a\}|} \sum_{\forall(i,j)\in a} w(e_i|f_j)   
\end{equation}

11) The sum of lexical translation probability differences between the human translation and the most probable translation according to the probability table. 
For each source word, we take the target word in human translation with the biggest probability. 
According to this method, we also count the unaligned words 
to calculate a ratio on the total number of tokens on each side. These features are calculated in the two directions of translation. 

We use the toolkit \textit{Scikit-Learn}~\cite{sklearn2011} to train different statistical machine learning classifiers.\footnote{The code and data set is publicly available at https://github.com/YumingZHAI/ctp.} 


\subsection{End-to-end Neural Network Architectures}

The source and target phrases are encoded using a bidirectional encoder with \gls{gru} (size $10$). The outputs of forward and backward recurrent networks are concatenated to form the source and target phrase representations (size $20$). After the encoder layer we have tried two different architectures. The first one is to build an alignment matrix for the source-target phrases, using the dot product of the two representations, inspired by these two work~\cite{Legrand2016,Pham2018}. Then a \textit{Convolutional Neural Network} (CNN) classifier is applied to this alignment matrix, which is composed of one convolution layer followed by pooling. Since the shape of the alignment matrix varies from one source-target pair to another, an adaptive pooling is used~\cite{pyramidpooling2015}. The output of the pooling layer is fed into a fully-connected layer followed by a linear layer as the output. In the second architecture, the source and target outputs of the encoder layer are averaged over time steps to produce two fixed-dimensional vectors, which are further concatenated (size $40$) and fed into a \gls{mlp} classifier. The hidden layer of \gls{mlp} includes $10$ hidden units with tanh non-linearity.  

The length of our phrases is usually short, especially for word-for-word \textit{Literal} instances. In order to build a more robust alignment matrix and to avoid the out-of-vocabulary problem, we finally choose to use character embeddings. As shown in table~\ref{tab:neuralclf-results-binary}, for the embedding layer, we have tried respectively randomly initialized character embeddings (size $10$), and training our own word embeddings using skipgram model of \textit{FastText}~\cite{fasttext_2016} on a TED Talks corpus (around 3M tokens for both English and French), with a word-embedding size of $100$, minimum n-gram of~$3$, and maximum n-gram of~$6$.
All the models have been trained in $200$ epochs, with a learning rate of $0.0001$ using Adam optimizer and the minibatch size of $20$. Dropout has been applied to all layers except the output and embedding layers. 

\vspace{-1em}

\section{Experimental Results and Analysis}
\label{experiment}

The table~\ref{tab:neuralclf-results-binary} and~\ref{tab:neuralclf-results-multi} show the results of our classifiers using end-to-end neural network architectures, for binary classification (balanced distribution) and multi-class classification. For the binary classification, \textit{Non\_literal} (NL) class has in total 1127 instances, and 1127 \textit{Literal} (L) instances are randomly chosen. Besides the preprocessing steps of lowercasing and correcting minor spelling errors, for the neural classifiers, we also normalized the clitic forms to complete words (\textit{e.g. 're $\rightarrow$ are}), and normalized digits to letter form (\textit{e.g. 42 $\rightarrow$ four two}). The architecture using word embeddings and \gls{mlp} obtain better results and is faster than the other two architectures. However, the current data set is too small for neural architectures to produce satisfactory results.  

\vspace{-2em}

\begin{table}[!ht]
\parbox{.5\linewidth}{
\centering
\caption{Binary classification \\(balanced distribution)}
\begin{tabular}{l|l|c|c|c} 
    \hline
    \multicolumn{2}{c|}{Architecture}  & Accuracy & F1 (L) & F1 (NL) \\ \hline
    \multicolumn{5}{c}{Randomly initialized character embedding} \\ \hline
    \multicolumn{2}{c|}{CNN} & 59.99\%  & 0.60   & 0.60   \\  
    \multicolumn{2}{c|}{MLP} & 71.16\%  & 0.71   & 0.71  \\ \hline   
    \multicolumn{5}{c}{Pre-trained fasttext word embedding} \\ \hline
    \multicolumn{2}{c|}{MLP} & \textbf{71.25\%}  & 0.71   & 0.71  \\ 
    \hline 
\end{tabular} 
\label{tab:neuralclf-results-binary}
}
\hfill
\parbox{.5\linewidth}{
\centering
\caption{Multi-class classification \\(five non-literal classes)}
\begin{tabular}{l|l|c|c|c} 
    \hline
    \multicolumn{2}{c|}{Architecture}  & Accuracy & Micro-F1 & Macro-F1 \\ \hline
    \multicolumn{5}{c}{Randomly initialized character embedding} \\ \hline
    \multicolumn{2}{c|}{CNN} & 34.08\%  & 0.34 & 0.20     \\  
    \multicolumn{2}{c|}{MLP} & 40.74\%  & 0.41 & 0.34      \\ \hline   
    \multicolumn{5}{c}{Pre-trained fasttext word embedding} \\ \hline
    \multicolumn{2}{c|}{MLP} & \textbf{43.22\%}  & 0.43 & 0.34  \\  
    \hline 
\end{tabular}  
\label{tab:neuralclf-results-multi}
}
\end{table}

The number of all non-literal instances (1127) is only one third of \textit{Literal} instances (3771). Considering this important difference, for the statistical machine learning classifiers, we first evaluated them under these configurations:  

-~six classes (\textit{Literal, Equivalence, Generalization, Particularization, Modulation, Contain\_Transposition}). We first put all \textit{Literal} instances. Then to have an approximately balanced class distribution, we randomly take 200 instances for \textit{Literal}. 

-~two classes (\textit{Literal} and \textit{Non\_literal}), with three distributions (3:1, 2:1, 1:1). The distribution 3:1 is the natural distribution in the data set. The instances of \textit{Literal} have been extracted randomly for the last two distributions. 

-~five classes (only non-literal categories). 

For each configuration, we have tuned the hyperparameters of different classifiers. We evaluate them by five-fold cross-validation,\footnote{
StratifiedKFold is used for cross-validation, where the folds are made by preserving the percentage of samples for each class.} using the metrics such as the average accuracy of five folds, the micro average and macro average F1-score~\cite{Tsoumakas2011}. The \textit{DummyClassifier} is used as a baseline, which generates random predictions by respecting the distribution of training classes. 

First, we attempted a direct classification into six classes (see table~\ref{tab:recap-configs}). The best results by \textit{RandomForest} reflect the difficulty of the task in multi-classes. On the other hand, we observe the potential of our features on classifying the category \textit{Literal} when the number of instances increases. As a result, we decide to divide the problem: conduct first a binary classification, and secondly a multi-class classification among the non-literal categories. 

\vspace{-0.5cm}

\begin{table}[!ht]
    \centering
    \caption{Classification results under different configurations, using all features}
    \begin{tabular}{l|c|c|c|c}
        \hline
        Distribution of classes  & Classifier                           &  Accuracy     & Micro-F1    & Macro-F1 \\ \hline
        \multicolumn{5}{l}{\textbf{Six classes}}  \\ \hline
         \multirow{2}{*}{six classes, with 3771 \textit{Literal}} & Dummy &  60.76\%     & 0.61        & 0.15    \\
                                                          & RandomForest &  \textbf{83.10\%}     & 0.83        & 0.44    \\ \hline
                                                          
        \multirow{2}{*}{six classes, with 200 \textit{Literal}} & Dummy &   18.92\%     & 0.19        & 0.16    \\       
                                                          & RandomForest &  57.04\%     & 0.57        & 0.52    \\ \hline 
        \multicolumn{5}{l}{\textbf{Two classes}}  \\ \hline
        \multirow{2}{*}{\textit{Literal} (3) : \textit{Non\_literal} (1)} & Dummy        &  65.84\%     & 0.66  & 0.52    \\                                                                                               & RandomForest &  \textbf{90.16\%}     & 0.90  & 0.86 \\ \hline 
        \multirow{2}{*}{\textit{Literal} (2) : \textit{Non\_literal} (1)} & Dummy        &  56.43\%     & 0.56  & 0.51    \\
                                                                          & RandomForest &  88.85\%     & 0.89  & 0.88    \\ \hline
        \multirow{2}{*}{\textit{Literal} (1) : \textit{Non\_literal} (1)} & Dummy        &  53.19\%     & 0.53  & 0.53    \\       
                                                                          & RandomForest &  \textbf{87.09\%}     & 0.87  & 0.87    \\ 
         \hline
        \multicolumn{5}{l}{\textbf{Five classes}}  \\ \hline
        \multirow{2}{*}{Five non-literal classes}                         & Dummy        & 20.32\%      & 0.20  & 0.18 \\
                                                                          & RandomForest & \textbf{55.10\%}      & 0.55  & 0.47  \\ 
        \hline 
    \end{tabular}
    \label{tab:recap-configs}
\end{table}

For the binary classification, the two best classifiers are \textit{RandomForest} and \gls{mlp}. Furthermore, \textit{RandomForest} has better performance than the two combined by the method \textit{hard voting} or \textit{soft voting}. The table~\ref{tab:recap-configs} presents the results under three different class distributions. From the natural distribution (3:1) to our artificial balanced distribution by randomly choosing \textit{Literal} instances (thus both class have 1127 instances), the average F1-score for the class \textit{Non\_literal} increases from 0.78 to 0.88. We will continue to test this tendency when a larger data set is available. Table~\ref{tab:recap-configs} also shows the results for the classification into five non-literal classes using all features, and the average F1-score for each non-literal category are shown in table~\ref{tab:5classes-eachClass}. The category \textit{Generalization} has many fewer instances than the other categories, which need to be augmented; there exist many confusions between \textit{Modulation} and the other categories, which suggests rather a review of annotation guidelines. 

\vspace*{-0.8cm}

\begin{table}[!ht]
    \centering
    \caption{Average F1-score for each non-literal class, using all features}
    \begin{tabular}{l|c|c|c|c|c}
    \hline
    Category & Equivalence & Generalization & Particularization & Modulation & Contain\_Transposition  \\ \hline 
    Nb. instances & 289       & 86           & 215             & 195      & 342               \\ \hline
    Average F1 & 0.51 & 0.25 & 0.56 & 0.36 & 0.68  \\ \hline 
    \end{tabular}
    \label{tab:5classes-eachClass}
\end{table}

\vspace*{-0.3cm}

The table~\ref{tab:feature-ablation} recapitulates the best performance on binary classification (balanced distribution) and on the classification of five non-literal classes, using the most helpful set of features. 
With the best performing classifier \textit{RandomForest}, we have investigated the performance of features one by one and also grouped them: \textit{PoS\_tagging} (feature 1, 2), \textit{surface} (feature 3), \textit{syntactic\_analysis} (feature 4, 5), \textit{external\_resource} (feature 6, 7, 8) and \textit{word\_alignment} (feature 9, 10, 11). For binary classification, feature 10 (bidirectional lexical weighting) is most helpful, which generates average F1-score of 0.78 for \textit{Literal} and 0.80 for \textit{Non\_literal} by itself. The group of features \textit{word\_alignment} contributes the most for the binary classification. The combination of all features generates the best results, which remain the same if we remove the feature 4 (constituency parsing), 7 (how the pair is linked in the resource \textit{ConceptNet}) and the features on \gls{pos} tagging apart from the vector counting the occurrence of each tag. The features in float form generally perform better than those in discrete form (\textit{e.g.} 0, 1, \textit{etc.}). Concerning the classification into five non-literal classes, the combination of all features except the group \textit{external\_resource} leads to the best results, where the group  \textit{PoS\_tagging} and \textit{syntactic\_analysis} contribute more than the group \textit{word\_alignment} and \textit{surface}. The accuracy changes from 55.10\% to 55.20\% after feature ablation (see table~\ref{tab:recap-configs}).

Our error analysis shows that in binary classification, it is difficult to distinguish \textit{Literal} and \textit{Equivalence}; in multi-class classification, the biggest confusion is between \textit{Equivalence} and \textit{Contain\_Transposition}. Consequently, we conducted another three binary classification experiments (see table~\ref{tab:groupClass}), where in all configurations each class has 549 instances to make the results comparable: i) \textit{Literal} vs \textit{Non\_literal} ii) \textit{Literal} combined with \textit{Equivalence} (E), vs the other classes iii) \textit{Literal} combined with \textit{Equivalence} and \textit{Transposition} (T), vs the other classes. The third configuration is more interesting, because the group of translation processes \textit{LET} do not bring meaning changes, while the processes \textit{non-LET} could. The results show that by including \textit{Transposition} (change grammatical classes without changing the meaning), the performance gets better than only grouping \textit{Literal} and \textit{Equivalence}, since we avoid the confusion between \textit{Equivalence} and \textit{Transposition}. The better results of binary classification (L vs NL, LET vs non-LET) indicate that in future work we can develop cascading classifiers, namely first separating word-for-word literal translations, or those which do not cause meaning changes, then conducting a finer-grained classification among the other categories. 


\begin{table}[!ht]
    \centering
    \caption{Classification results after feature ablation study}
    \begin{tabular}{L{3cm}|c|c|c}
    \hline
     & average accuracy & \multicolumn{2}{c}{average F1-scores} \\  \hline
    binary classification (balanced distribution) & \textbf{87.09\%} & 0.87 (Literal) & 0.88 (Non\_literal)  \\ \hline 
    five non-literal classes & \textbf{55.20\%} & 0.55 (micro average) & 0.48 (macro average) \\ \hline
    \end{tabular}
    \label{tab:feature-ablation}
\end{table}

\begin{table}[!h]
    \centering
    \caption{Classification results after grouping classes, every class has 549 instances}
    \begin{tabular}{c|c|c|c}
     \hline
     Configuration & average accuracy & average F1 (class1) & average F1 (class2)  \\ \hline
     Dummy   & 48.63\% & 0.49 & 0.49 \\ \hline
     L vs NL & 85.24\% & 0.84 & 0.86  \\ 
     LE vs non-LE & 75.32\% & 0.74 & 0.77 \\ 
     LET vs non-LET & 79.42\% & 0.78 & 0.81 \\ \hline  
    \end{tabular}
    \label{tab:groupClass}
\end{table}


\section{Conclusion and Perspectives}

We have proposed a new Natural Language Processing task of automatically classifying translation processes at subsentential level, based on manually annotated examples from a parallel English-French TED Talks corpus. 
To the best of our knowledge, these translation processes have not been explicitly exploited during paraphrase extraction from bilingual parallel corpora.
With the best performing classifier \textit{RandomForest} and feature engineering, our empirical results show a best accuracy of 87.09\% for binary classification (\textit{Literal} vs \textit{Non\_literal}) and 55.20\% for multi-class classification (\textit{Equivalence, Generalization, Particularization, Modulation, Contain\_Transposition}), which are much better than the baseline random classifier. 

This task is complicated, and our exploratory work is restrained by the limited amount of annotated examples. However, our work demonstrates that automatically classifying translation processes seem possible, and the experiments show the directions that we can follow in future work. 
There is much room to constitute an augmented and balanced data set, on which we will evaluate our classifier to observe the performance. The finer error analysis of the classification results is useful to help the research on corpus annotation and linguistic analysis. 
We will continue to improve the classifier on English-French, by implementing other features for multi-class classification, and explore more neural architectures. We will also extend our work to English-Chinese translation pairs. 
One of our long term objectives is leveraging this automatic classification to better control paraphrase extraction from bilingual parallel corpora. 


\vspace{-1em}

\bibliographystyle{splncs04}
\bibliography{cicling}

\end{document}